\pdfoutput=1
\documentclass[11pt]{article}
\usepackage{acl}
\usepackage{times}
\usepackage{latexsym}
\usepackage[T1]{fontenc}
\usepackage[utf8]{inputenc}
\usepackage{microtype}
\usepackage{amsmath}
\usepackage{braket}
\usepackage{subfigure}
\usepackage[export]{adjustbox}
\usepackage{adjustbox}
\usepackage{graphicx}
\usepackage{url}
\usepackage{bm}
\usepackage{multirow}
\usepackage{tikz}
\usepackage[ruled,vlined,algo2e]{algorithm2e}
\usepackage{algorithm}
\usepackage{algorithmic}
\usepackage{graphicx}
\usepackage{setspace}
\usepackage{amsmath}
\usepackage{hyperref}
\usepackage{enumitem}
\usepackage{CJKutf8}
\usepackage{float}
\usepackage{makecell}

\usepackage{booktabs}
\usepackage{multirow}
\usepackage{subfigure}

\let\Algorithm\algorithm
\renewcommand\algorithm[1][]{\Algorithm[#1]\setstretch{1}}

\usepackage{xcolor}
\usepackage[ruled,vlined,algo2e]{algorithm2e}

\SetCommentSty{mycommfont}
\AtBeginDocument{
  \let\mathbb\relax
  \DeclareMathAlphabet{\mathbb}{U}{msb}{m}{n}
}

\usepackage{amssymb}
\usepackage{pifont}
\title{MCML: A Novel Memory-based Contrastive Meta-Learning Method \\ for Few Shot Slot Tagging}

\author{Hongru Wang, Zezhong Wang, Wai Chung Kwan, Kam-Fai Wong \\
MoE Key Laboratory of High Confidence Software Technologies \\
The Chinese University of Hong Kong \\
\{hrwang, kfwong\}@se.cuhk.edu.hk
}
\date{}

\begin{document}
\maketitle

\begin{abstract}
Meta-learning is widely used for few-shot slot tagging in task of few-shot learning. The performance of existing methods is, however, seriously affected by \textit{sample forgetting issue}, where the model forgets the historically learned meta-training tasks while solely relying on support sets when adapting to new tasks. To overcome this predicament, we propose the \textbf{M}emory-based \textbf{C}ontrastive \textbf{M}eta-\textbf{L}earning (aka, MCML) method, including \textit{learn-from-the-memory} and \textit{adaption-from-the-memory} modules, which bridge the distribution gap between training episodes and between training and testing respectively. Specifically, the former uses an explicit memory bank to keep track of the label representations of previously trained episodes, with a contrastive constraint between the label representations in the current episode with the historical ones stored in the memory. In addition, the \emph{adaption-from-memory} mechanism is introduced to learn more accurate and robust representations based on the shift between the same labels embedded in the testing episodes and memory. Experimental results show that the MCML outperforms several state-of-the-art methods on both SNIPS and NER datasets and demonstrates strong scalability with consistent improvement when the number of shots gets more. 
\end{abstract}

\section{Introduction}



\begin{figure*}[t]
\centering
\includegraphics[trim={16cm 14cm 17cm 4cm}, clip, width=1.0\textwidth]{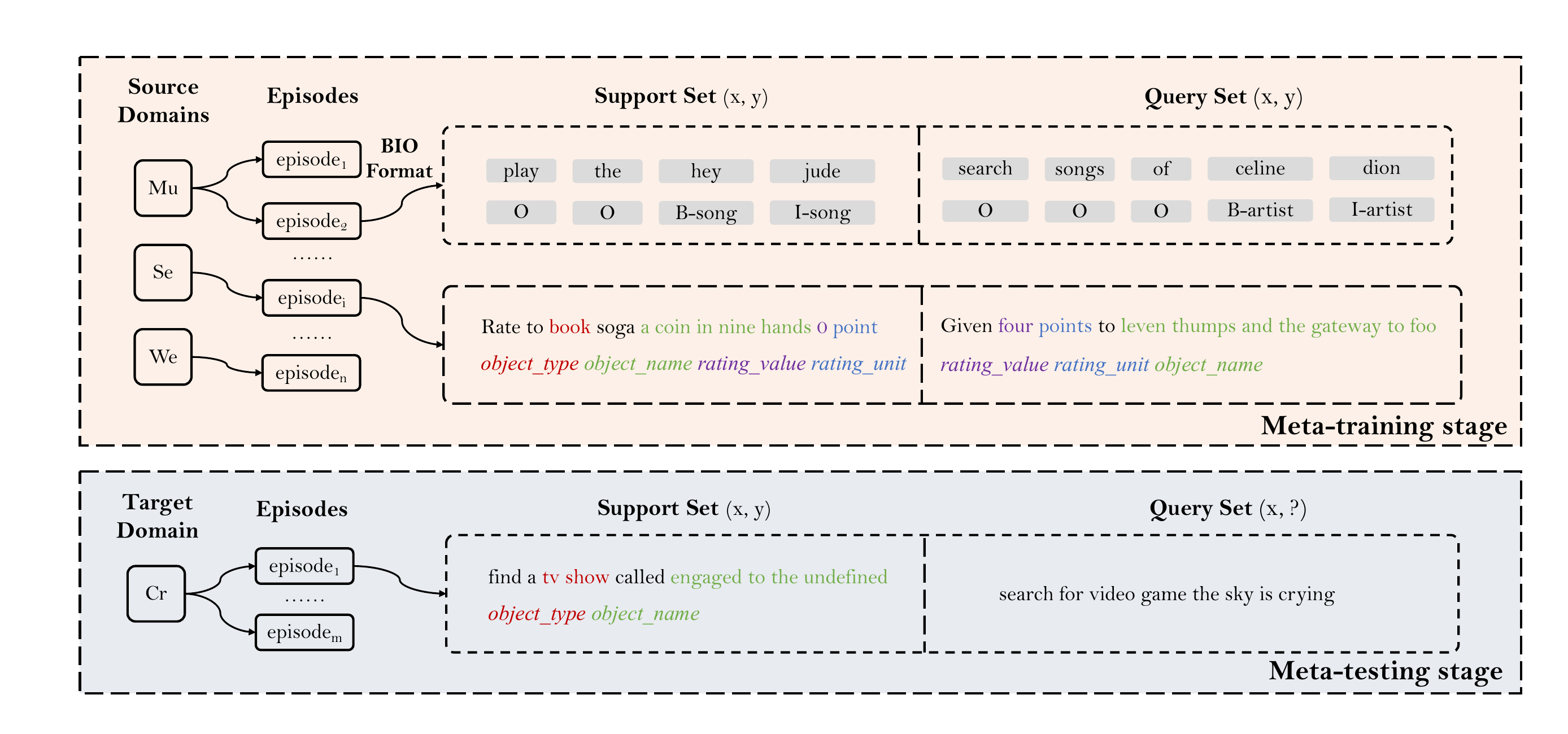}
\caption{Episode-setting of Metric-based meta-learning, where each domain contains multiple episodes and each episode consists of a support-query-set pair. Different colors indicate different labels.}
\label{fig:episde_setting}
\end{figure*}

Slot tagging \cite{tur2011spoken}, is a key part of natural language understanding, which is usually modeled as a sequence labeling problem with BIO format as shown in Figure~\ref{fig:episde_setting} \cite{chen2019bert}. However, rapid domain transfer and scarce labeled data in the target domain introduce new challenges \cite{bapna2017towards,zhang-etal-2020-mzet}. To this end, significant efforts have been made to develop few-shot techniques \cite{1597116,snell2017prototypical,vinyals2017matching}, which aim to recognize a set of novel classes with only a few labeled samples (i.e, less than 50-shot) by knowledge transfer from a set of base classes with abundant annotated samples.

Among several few-shot learning approaches \cite{9428530}, metric-based meta-learning has been widely used in slot tagging because they are model-agnostic, effective, and easily applicable \cite{snell2017prototypical,vinyals2017matching,zhu2020vector,hou2020fewshot}. To cope with the data scarcity of novel classes, metric-based methods split data into different episodes deliberately while each episode consists of one support set and one query set. The model classifies a (query) item according to its similarity with the representation of each label learned from the support set in this episode.

However, this kind of setting has shown several limitations, where the similarity calculation conducted only at the episode level hinders the learning of the original representations, resulting in the \textit{sample forgetting problem} \cite{toneva2018empirical}. On the one hand, this cripples the model’s ability to learn consistent representations for the same labels across different episodes. Here, the same labels may occur during different episodes at the meta-training stage and also possibly span the meta-training and meta-testing stages. For example, \textit{B-Object\_name} occurs in both the meta-training stage and meta-testing stage as shown in Figure~\ref{fig:episde_setting}. On the other hand, the similarity calculation is conducted between the query and support set only in one episode under the few-shot setting, resulting in the representation shift while ignoring the same label representation in previous episodes. First of all, the representation of each label is not accurate due to the limited labeled samples. Besides that, the locally closest label in one episode is not necessarily the globally closest. Furthermore, with the number of shots increasing, the sample forgetting problem becomes worse and the model performance saturates quickly \cite{cao2019theoretical}. 

To overcome the above limitations, we propose the Memory-based Contrastive Meta-learning (aka, MCML) method, marrying the benefits of \textit{learn-from-the memory} and \textit{adaption-from-the-memory} to capture more transferable and informative label representations. Specifically, during meta-training, we use an explicit memory bank to keep track of the label representations from the historical episodes. Then a contrastive constraint is added to pull together semantically similar (i.e, positive) samples in the embedding space while pushing apart dissimilar (i.e, negative) samples. This is what we call the \textit{learn-from-the-memory} technique. Secondly, during meta-testing, we use the \textit{adaption-from-the-memory} technique to bridge the shift between the input labels embedded in the test episodes and the label anchors in the memory. In addition, an indicator is used to control how much information we want to acquire from the memory. The combination of \textit{learn-from-the-memory} and \textit{adaption-from-the-memory} helps the model to learn consistent representations for the same labels and distinguished representations for different labels concurrently across different episodes. To summarize, our contributions are three-fold: 

\begin{itemize}
    \item This is the first work to tackle the \textit{sample forgetting problem} of metric-based methods. We propose a novel Memory-based Contrastive Meta-learning (MCML) method to bridge the gap between different episodes.
    \item We propose two model-agnostic methods including learn-from-the-memory and adaption-from-the-memory, which can be applied in different stages separately. The combination of them achieves the best performance even with the number of shots increasing. 
    \item The experimental results confirm the effectiveness of our model with very favorable performance over several state-of-the-art methods on both \textit{SNIPS} and \textit{NER} datasets.
\end{itemize}

\section{Related Work}

Few-shot learning was first proposed as a transfer method using a Bayesian approach on low-level visual features \cite{1597116}. Over the past few years, researchers have developed alternative techniques to build domain-specific modules for low-resource cross-domain natural language understanding \cite{bapna2017zeroshot,lee2018zeroshot,Fritzler_2019,shah2019robust}. Most recent works have tried to model the transition possibility or similarity function between different labels with the metric-based meta-learning framework as backbone \cite{hou2020fewshot,zhu2020vector,wang-etal-2022-prior}. Nevertheless, episode-level relationships are still under-explored in previous works, except for a number of methods on image classification \cite{li2019lgm,sun2019meta,ouali2020spatial,fei2021melr}. \citet{fei2021melr} proposed a novel method to learn more robust representations by sampling two episodes containing the same set of classes for meta-training while \citet{ouali2020spatial} used intra-episode spatial contrastive learning (SCL) as an auxiliary pre-training objective to learn general-purpose visual embeddings for image classification.

Distinguishing from prior work, we first exploit the inter-episode relationship for natural language understanding by using an explicit memory bank. Most researchers choose to store the encoded contextual information in each meta episode under the few-shot setting \cite{kaiser2017learning,cai2018memory}. Another alternative method adopts parameterized memory network to implicitly save historical information \cite{geng2020dynamic}. Our work also keeps in line with Momentum Contrast(MoCo) which utilizes an external memory module to store positive or negative samples for contrastive learning \cite{he2020momentum}. Similarly, with a relatively large size of samples, unilateral representations from one episode in the metric-based methods can be alleviated.

\begin{figure*}[t]
\centering
\includegraphics[trim={2cm 5cm 0cm 4cm}, clip, scale=1.0, width=1.0\textwidth]{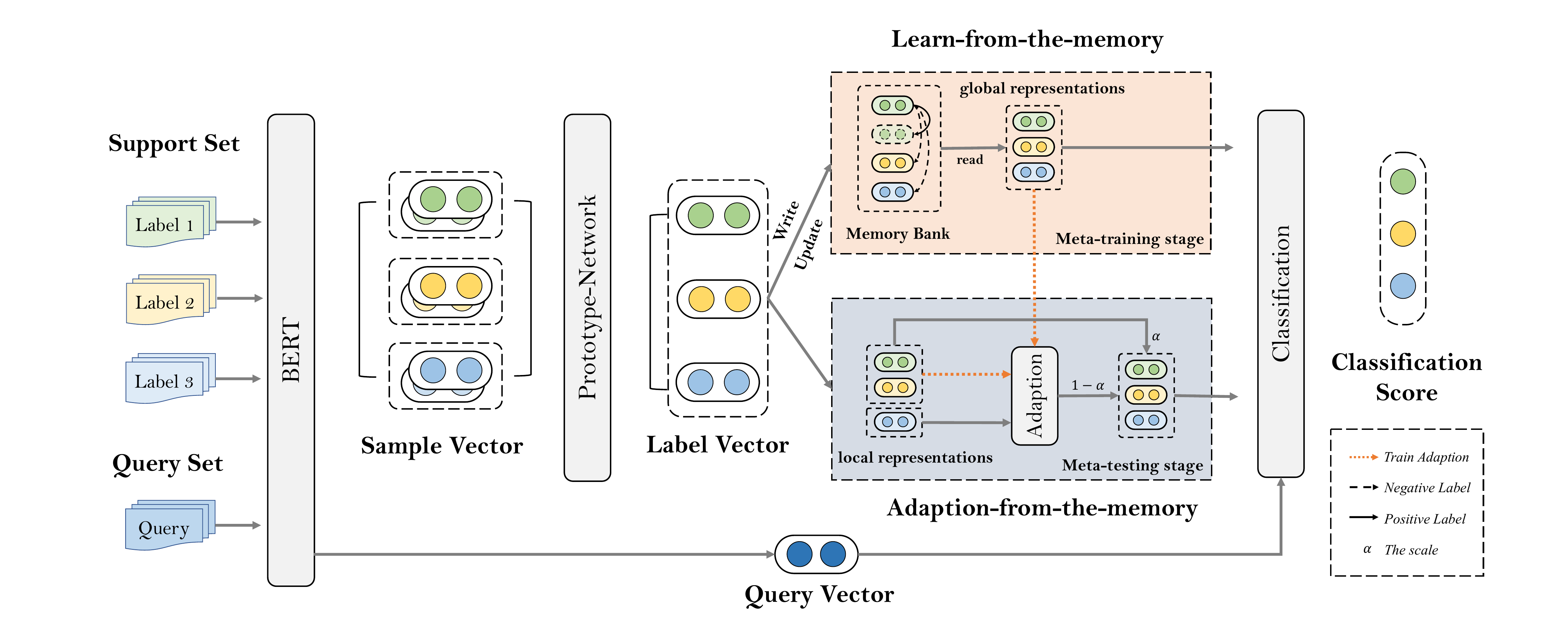}
\caption{The overview of Our Framework, including three modules 1) Memory Bank, 2) Learn-from-the-memory, and 3) Adaption-from-the-memory, while the memory bank stores global representations across different episodes, learn-from-the-memory is used to learn more accurate and robust representations during the meta-training stage, and adaption-from-the-memory to bridge the gap between training and testing for local representations in test episode. }
\label{model}
\end{figure*}

\section{Preliminaries}
Before introducing our proposed framework, we provide the problem definition and an illustration of the basic framework of metric-based meta-learning to solve few-shot slot tagging in this section. 

\subsection{Problem Definition}
We denote each sentence $x = (x_1, x_2, x_3, ..., x_p )$ and the corresponding label $y = (y_1, y_2, y_3, ..., y_p)$. Usually, we are provided with lots of labeled data (i.e. $(x, y)$ pairs) of source domains $D_s$, and few-shot (less than 50) labeled data as well as plenty of unlabeled data in the target domain $D_t$ under the few shot setting. We split the data as episodes $e = (S, Q)$ in which $S = \{ x_i^j, y_i^j\}_{i = 1 ... N}^{j = 1 ... K}$ and $Q = \{x^j, y^j\}_{j=1}^{|Q|}$ respectively. $S$, as the support set, contains $K$ examples (K-shot) for each of $N$ labels (N-way) while $Q$ contains several unlabeled samples \footnote{We have labels during meta-training.}. Thus, the few-shot model is trained based on many episodes $E_{tr} = (e_1, e_2, e_3, ..., e_n)$ initially. The trained model is then directly evaluated on the target domain $E_{te} = (e_1, e_2, ..., e_m)$. The objective is formulated as follows:

\vspace{-4mm}
\begin{equation}
    y^* = \operatorname{argmax}_y p_\theta (y|x,S)
\end{equation}

\noindent where $\theta$ refers the parameters of the slot tagging mode, the $(x,y)$ pair and the support set from the target domain, $E_{tr}$ and $E_{te}$ represent different episodes during meta-training and meta-testing respectively.

\subsection{Metric-based Meta Learning}
Given an episode consisting of a support-query set pair, the basic idea of metric-based meta-learning \cite{zhu2020vector,hou2020fewshot} is to classify an item (a sentence or token) in the query set based on its similarity with the representation of each label, which is learned from the few labeled data of the support set. Some representative works are matching network \cite{vinyals2017matching} and prototypical network \cite{snell2017prototypical}. More specifically, given an input episode $(S, Q)$ pair, the model encodes these two parts to get the sample vector and query vector respectively:

\vspace{-4mm}
\begin{equation}
    \bm{S}, \bm{Q} = Encoder(S, Q)
\end{equation}

After that, various models can be used to extract label representations $c_{y_i}$. Take the prototypical network as an example, each prototype (label representation) is defined as the average vector of the embedded samples which have the same labels:

\vspace{-4mm}
\begin{equation}
    c_{y_i} = \frac{1}{N_{y_i}} \sum_{i=1}^{N} \sum_{j=1}^{K} I\{y_i^j = y_i\} s_i^j
\end{equation}

while $I$ is an indicator function which equals to True when $y_i^j == y_i$ else False; $s_i^j$ is the corresponding sample vector from $\bm{S}$. 

Lastly, we calculate the distance between the label representation and the sample vector from the query set. The most popular distance function is the dot product function which is defined as follows:

\vspace{-4mm}
\begin{equation}
    SIM(x_i, c_k) = x_i^T c_k 
\end{equation}

\vspace{-4mm}
\begin{equation}
    y = Softmax ( SIM(x_i, c_k) ) 
\end{equation}

The label of instance (i.e, $x_i$) from the query set is the label whose embedding is closest with the instance vector (i.e, $c_i$). This can be calculated through a softmax layer. However, in this way, the learned prototype of labels may lack general discriminative semantic features since it ($c_i$) only needs to be closer to instance ($x_i$) compared with other prototypes (i.e, $c_j$ where j!=i) in the same episode without considering the global prototypes. Since the support set may only contain a few instances with the same label, the representation becomes imprecise and fragile \cite{ouali2020spatial}. 

\section{Model}
In this section, we first illustrate the overview of our proposed framework (Section \ref{framework}), and then we discuss how to learn and adaption from memory (Section \ref{learn} and \ref{adaption}) respectively.

\subsection{Framework}
\label{framework}

Due to data scarcity and domain transfer, \textit{sample forgetting problem} seriously hinders the model to learn robust representation, resulting in worse adaptability. To overcome this problem during meta-training and meta-testing stages, we propose \textit{learn-from-the-memory} and \textit{adaption-from-the-memory} techniques respectively as shown in Figure~\ref{model} to reuse the learned representations \cite{raghu2019rapid}.

\noindent \textbf{Learn-from-the-memory:} During the meta-training stage, the model will continuously train on different episodes. We utilize an external memory bank to store all learned label representations from the support set. These representations form different clusters naturally according to their original labels. When a newly seen label appears, a contrastive loss is computed on these dimensional representations by attracting positive samples, which have the same label, and by repelling the negative samples which have different labels. If this label has not been encountered before, we just write it into our memory.

\noindent \textbf{Adaption-from-the-memory:} During the meta-testing stage, we first learn an adaption layer by using these overlapped labels during meta-training and meta-testing, and then we use the learned adaption layer to project these unseen (i.e, not overlap) labels from testing space to training space in order to bridge the shift between testing space and training space. In addition, we use the skip connection to control how much information we want to acquire from the memory.

\subsection{Learn from Memory}
\label{learn}

To consider all prototypes and learn consistent representations during meta-training, we first use a memory bank to store all prototypes of different labels from the support set. We design three basic operations in the memory bank: \textit{write, update,} and \textit{read}. 

\noindent \textbf{Write. }Specifically, starting from the first episode $e_1$ to the last episode $e_n$ in $E_{tr}$, we store the label representations from the corresponding support set $C_i = (c_1, c_2, ... c_k)$ into external memory bank $M$ with the label name as key, where $k$ is the number of labels for the current episode. $M$ increases as the episode continue on. 

\vspace{-4mm}
\begin{equation}
    \Bar{k} \leq M \leq m * \Bar{k}
\end{equation}

while $\Bar{k}$ represents average number of labels for all episodes, and $m$ is the number of episodes. For the $i$th episode, we first calculate the prototypical embedding of seen-label clusters from memory. Theoretically, this step is unnecessary but we choose to do so to save computational resources\footnote{
We call the prototypical embedding of label clusters by centroid (i.e, prototype in historical episodes), and prototype as average label embedding in one episode. Here if we skip the calculation of centroid, then we can directly use these prototypes.}.

\begin{equation}
\label{centroid}
    c_k^{\star} =  \frac{1}{N_k} \sum_{i=1}^{N_k} I\{c_i = c_k\} c_i
\end{equation}

We use $c_k^{\star}$ to represent the centroid of the $k$th cluster and we also store it in the memory bank, and then we define a distance function following \cite{ding2021prototypical} as follows:

\vspace{-4mm}
\begin{equation}
    d (c_i, c_j) = 1 / (1 + \exp( \frac{c_i}{||c_i||} \cdot \frac{c_j}{||c_j||})
\end{equation}

\textbf{Read. } For the coming labels and corresponding representations, there are two situations: (1) the label is new which means it never appears in the previous episodes, and (2) the label has already been stored in the memory bank. We extract all centroid representations from the memory bank and impose a contrastive learning constraint accordingly.

\vspace{-4mm}
\begin{equation}
\begin{split}
    L_{memory} = - \frac{1}{K} \sum_{c_i \in S^c, c_j \in S^{-c}} [\log d(c_i, c_k^{\star})  \\
    + \log(1- d(c_j, c_k^{\star}))]
\end{split}
\end{equation}

For the new label, there are no positive pairs, and we increase the distances between its representation and all extracted representations. For the same label, we draw the same centroid representation but repel different centroids.

This objective effectively serves as regularization to learn more consistent and transferable label representation as they evolve during meta-training \cite{ding2021prototypical,he2020momentum}. We emphasize that the parameters of models do not change at this stage, and we do not need to modify the architecture of traditional metric-based meta-learning models. As such, the model can be easily trained together with other components in an end-to-end fashion.

\begin{figure}[t]
\centering
\includegraphics[trim={30cm 14cm 17cm 18cm}, clip, width=1.0\textwidth]{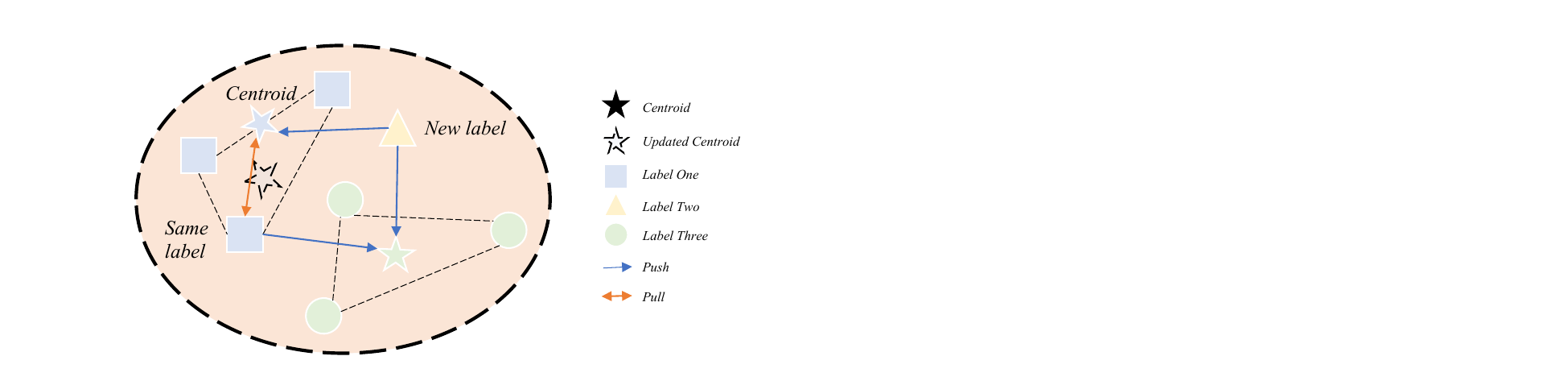}
\caption{The high-level overview of our \textit{learn-from-the-memory}. There are multiple label representations in the memory bank. We will attract representations with the same label, repel otherwise, and also update accordingly.}
\label{fig:learn}
\end{figure}


\textbf{Update. }At the last, we need to re-calculate the prototypical embeddings of seen-label clusters in the memory following Equation \ref{centroid}. In this way, the distribution shift across different episodes during meta-training will be alleviated and thus more general discriminative representations can be learned. Figure \ref{fig:learn} demonstrates the whole processing of \textit{learn-from-the-memory}.

\subsection{Adaption from Memory}
\label{adaption}

To address the forgetting problem during the meta-testing stage, we take advantage of stored representations in the memory bank to build a bridge connecting the testing space and training space. With overlapped labels between meta-training and meta-testing, two types of representations can be observed: 1) one from memory during meta-training; 2) the other from the current episode during meta-testing. It is noted that labels overlap frequently in practice, e.g. \textit{B-person} and \textit{B-city} almost appear in every slot tagging dataset. 

\begin{figure}[t]
\centering
\includegraphics[trim={29cm 12cm 10cm 15cm}, clip, width=1.0\textwidth]{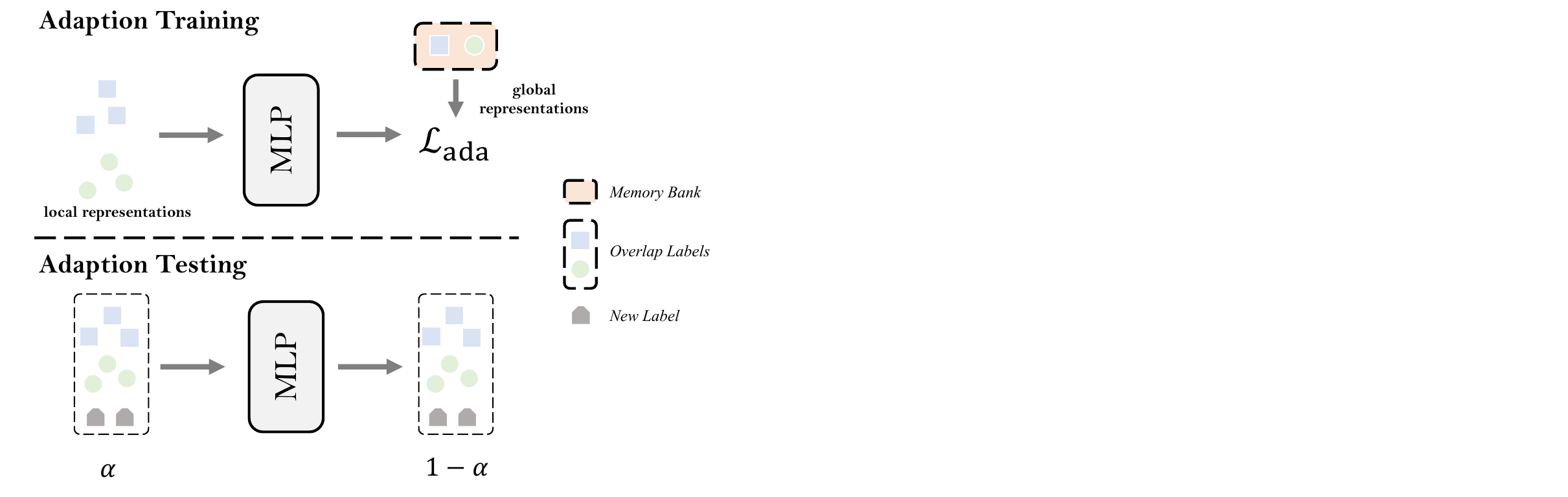}
\caption{The high-level overview of our \textit{adaption-from-memory}. We extract global representations from the memory bank which stores label representations during the meta-training stage and use them as grounded truth to learn the adaption layer (aka, MLP) with label representations from the current episode (aka, local representations) as input. We pass all labels, including overlap labels and new labels, from the current episode to the adaption layer to get the final representations.}
\label{fig:adaption}
\end{figure}

As shown in Figure \ref{fig:adaption}, we decompose the whole process into two steps. First of all, we use the overlapped labels during meta-training and meta-testing to learn the adaption function $f$ which minimizes the representation gap between meta-training and meta-testing \footnote{We emphasize this operation is conducted at episode-level to comply with the few-shot setting.}.

\vspace{-4mm}
\begin{equation}
    y_i = f ( R_{test\_overlap}^i )
\end{equation}

Where $f$ can be implemented by Multilayer Perceptron (MLP) or one linear layer with the following loss function. Here $R_{test\_overlap}^i$ means $i$th label appears in both the memory bank and current test episode. The learning objective of adaption layers $L_{ada}$ can be defined as follows:

\vspace{-6mm}
\begin{equation}
    L_{ada} = \sum_{i=1}^{|Overlap|} || R_{train\_overlap}^i - y_i ||^2
\end{equation}

Where $R_{train\_overlap}^i$ here can be directly extracted from the memory bank as the ground truth representation. We then use the learned adaption function to project the new labels (i.e, not overlap) to the training space based on the assumption that the training space should be more accurate than the testing space which consists of more labeled data.

\vspace{-4mm}
\begin{equation}
    R_{train\_new} = f ( R_{test\_new} )
\end{equation}

In this case, we can get original representations in the testing space and representations in the training space after adaption for both overlap labels and new labels. Our final representation for each label can be the combination of these two kinds of representation from different spaces.

\vspace{-6mm}
\begin{equation}
     R_{fin} = \alpha * R_{ori} + (1 - \alpha) * R_{ada}
\end{equation}

where $\alpha \in (0, 1)$ is a hyper-parameter that controls the percentage of information from the original testing space and from adaption. By adaption from the memory, the distribution shift in the testing episodes rooting in domain transfer and few-shot setting will be de-biased by the global representations in the memory.

\begin{table*}[h]
\centering
\begin{tabular}{l|l|lllllll|l}
\toprule
\multirow{2}{*}{\textbf{N-shot}} & \multirow{2}{*}{\textbf{Model}} & \multicolumn{7}{c|}{\textbf{Domain}}  & \multirow{2}{*}{\textbf{Avg.}}  \\ 
& & We & Mu & Pl & Bo & Se & Re & Cr & \\
\midrule
\multirow{6}{*}{\textbf{1-shot}} & SimBERT   & 36.10  & 37.08 & 35.11 & 68.09  & 41.61 & 42.82   & 23.91 & 40.67 \\
   & TransferBERT & 55.82  & 38.01  & 45.65  & 31.63  & 21.96    & 41.79  & 38.53 & 39.06  \\
  & L-TapNet+CDT+PWE   & 71.53   & \textbf{60.56}  & 66.27          & 84.54  & 76.27  & 70.79  & 62.89  & 70.41 \\
  & L-ProtoNet+CDT+VPB  & \textbf{73.08} & 58.50 & 68.81 & 82.41 & 75.88 & \textbf{73.17} & 70.27 & 71.73 \\
   & Coach \cite{liu-etal-2020-coach} &  55.81 & 38.72 & 41.60 & 41.44 & 35.25 & 54.38 & 47.74 & 44.99 \\
\cline{2-10}
&Ours & 72.30 & 58.33 & \textbf{69.64} & 82.90 & \textbf{77.23} & 72.79 & \textbf{79.57} & \textbf{73.25} \\
\midrule
\multirow{6}{*}{\textbf{5-shot}}  & SimBERT  & 53.46 & 54.13  & 42.81  & 75.54 & 57.10
& 55.30  & 32.38  & 52.96 \\
& TransferBERT & 59.41  & 42.00  & 46.07 & 20.74 & 28.20 & 67.75  & 58.61 & 46.11 \\
& L-TapNet+CDT+PWE  & 71.64 & 67.16  & 75.88   & 84.38       & 82.58          & 70.05       & 73.41  & 75.01 \\
& L-ProtoNet+CDT+VPB & \textbf{82.54} & 69.52 & 80.45 & 91.03 & 86.14 & 80.75 & 75.95 & 80.91 \\
\multicolumn{1}{c|}{}   & Coach \cite{liu-etal-2020-coach} & 73.56 & 45.85 & 47.23 & 61.61 & 65.82 & 69.99 & 57.28 & 60.19  \\
\cline{2-10}
&Ours & 81.79 & \textbf{69.70} &  \textbf{80.78} & \textbf{91.53} & \textbf{87.09} & \textbf{82.49} & \textbf{81.07} & \textbf{82.06}\\
\bottomrule
\end{tabular}
\caption{$F_1$ scores on few-shot slot tagging of the SNIPS dataset}
\label{tab:snip_main}
\end{table*}

\subsection{Training Objective}
The learning objective of our methods is the sum of three parts. It is noted that these losses are not optimized simultaneously.

\vspace{-4mm}
\begin{equation}
    L_{ner} = \sum_{c=1}^{K} y_{c} \log(P_c)
\end{equation}

\vspace{-4mm}
\begin{equation}
    L = L_{ner} + L_{memory} + L_{ada}
\end{equation}

while $L_{ner}$ represents the traditional cross-entropy loss of sequence labeling (see Eq. 14) and is optimized with $L_{memory}$ during training (see Eq. 9). $L_{ada}$ is optimized during testing (see Eq. 11).

\section{Experiments}
\subsection{Datasets}

We evaluate the proposed methods following the data split setting provided by \cite{hou2020fewshot} on NER and SNIPS datasets \cite{coucke2018snips}. It is in the episode data setting \cite{vinyals2017matching}, where each episode contains a support set (1-shot or 5-shot) and a batch of labeled samples. For NER, we followed the setting as same as \cite{zhu2020vector}, which contains 4 different datasets: CoNLL-2003 (i.e. News) \cite{tjong-kim-sang-de-meulder-2003-introduction}, GUM (i.e. Wiki) \cite{10.1007/s10579-016-9343-x}, WNUT-2017 (i.e. Social) \cite{derczynski-etal-2017-results} and OntoNotes (i.e. Mixed) \cite{pradhan-etal-2013-towards}. For SNIPS, it consists of 7 domains with different label sets: Weather (We), Music (Mu), PlayList (Pl), Book (Bo), Search Screen (Se), Restaurant (Re) and Creative Work (Cr). And also, we extend our method to more shots (10-shot and 20-shot) to further demonstrate the effectiveness and robust generalization capability of our approach.

\begin{table*}[h]
\centering
\resizebox{160mm}{18mm}{
\begin{tabular}{l|llll|l||llll|l}
\toprule[1pt]
\multirow{2}{*}{\textbf{Model}} & \multicolumn{5}{c||}{\textbf{1-shot}}  & \multicolumn{5}{c}{\textbf{5-shot}}   \\\cline{2-11} 
& News. & Wiki & Social & Mixed & \textbf{Avg.} & News & Wiki & Social & Mixed & \textbf{Avg.} \\
\midrule
SimBERT & 19.22 & 6.91 & 5.18 & 13.99 & 11.32 & 32.01 & 10.63 & 8.20 & 21.14 & 18.00 \\
TransferBERT &  4.75 & 0.57 & 2.71 & 3.46 & 2.87 & 15.36 & 3.62 & 11.08 & 35.49 & 16.39 \\
L-TapNet+CDT+PWE & 44.30 & 12.04 & 20.80 & 15.17 & 23.08 & 45.35 & 11.65 & 23.30 & 20.95 & 25.31 \\
L-ProtoNet+CDT+VPB & 42.23 & 11.36 & \textbf{27.72} & \textbf{31.17} & 28.10 & 56.30 & 19.17 & 34.95 & \textbf{43.30} & 38.43 \\
\hline
Ours & \textbf{42.70} & \textbf{13.20} & 26.75 & 29.86 & \textbf{28.13} & \textbf{56.89} & \textbf{22.09} & \textbf{35.27} & 42.08 & \textbf{39.08} \\
\bottomrule[1pt]
\end{tabular}
}
\caption{$F_1$ scores on few-shot slot tagging of the NER dataset}
\label{tab:ner_main}
\end{table*}

\subsection{Baselines}
\noindent \textbf{SimBERT. } It assigns labels to words according to the cosine similarity of word embedding of a fixed BERT. For each word $x_i$, SimBERT ﬁnds the most similar word $x_k$ in the support set and assigns $x_k$'s label to $x_i$.

\noindent \textbf{TransferBERT. } It directly transfers the knowledge from the source domain to the target domain by parameter sharing. We train it on the source domain and select the best model on the same validation set of our model. Before evaluation, we ﬁne-tune it on the target domain support set.

\noindent \textbf{L-TapNet+CDT+PWE } \cite{hou2020fewshot} one of the strong baselines for few-shot slot tagging, which enhances the WarmProtoZero(WPZ) \cite{Fritzler_2019} model with label name representation and incorporate it into the proposed CRF framework.

\noindent \textbf{L-ProtoNet+CDT+VPB } \cite{zhu2020vector} current state-of-the-art metric-based meta-learning, which investigates the different distance functions and utilizes the distance function VPB to boost the performance of the model. 

\noindent \textbf{Coach} \cite{liu-etal-2020-coach} Coarse-to-fine approach (Coach) for cross-domain slot filling, which is a current state-of-the-art few-shot fine-tuning method incorporating template regular loss and slot description information.

\subsection{Implementation Details}
We take the pre-trained uncased \verb|BERT-Base| \cite{devlin2019bert} as an encoder to embed words into contextually related vectors in all experiments. Following the setting in \cite{zhu2020vector}, we use \verb|ADAM| \cite{kingma2014adam} to train the model with a learning rate of 1e-5, a weight decay of 5e-5. And we set the distance function as \verb|VPB| \cite{zhu2020vector}. To prevent the impact of randomness, we test each experiment 10 times with different random seeds following \cite{hou2020fewshot}. For adaption from memory, we set the iteration as 1000, and $\alpha$ from $[0.1, 0.3, 0.5, 0.7, 0.9]$ and report the best result.

\subsection{Main Result}
Table~\ref{tab:snip_main} and Table~\ref{tab:ner_main} show the results of both 1-shot and 5-shot slot tagging of SNIPS and NER datasets respectively. Our method reaches comparable results with the state-of-the-art and outperforms in 3 out of 7 domains under 1-shot setting, and 6 under 5-shot setting at SNIPS dataset. Specifically, our method achieves about 13\% ($70.27 \rightarrow 79.57$) and 7\% ($75.95 \rightarrow 81.07$) improvements in the Cr domain under 1-shot and 5-shot respectively. Besides that, the improvement keeps consistent on the NER dataset while adding additional shots leads to greater improvement. It's obvious that our method demonstrates strong scalability and flexibility with the number of shots increasing. When comparing Coach \cite{liu-etal-2020-coach} with L-TapNet+CDT+PWE \cite{hou2020fewshot} and L-TapNet+CDT+VPB \cite{zhu2020vector}, it is also interesting to see that fine-tuning is not as competitive as metric-based approaches when the shot is smaller.

\begin{table*}[h]
\centering
\begin{adjustbox}{max width=1.0\textwidth}
\begin{tabular}[scale=1.0]{l|ccc|ccc|ccc|ccc}
\toprule[1pt]
  & \multicolumn{3}{|c|}{\textbf{1-shot}} & \multicolumn{3}{|c}{\textbf{5-shot}} & \multicolumn{3}{|c|}{\textbf{10-shot}}
  & \multicolumn{3}{|c}{\textbf{20-shot}}  \\
\cline{2-13}
  & B & A & M & B  & A & M & B & A & M  & B & A & M  \\
\midrule
\textbf{We} & 73.08 & 72.30 & 71.83 & 82.54 & 81.13 & 81.79  & 79.09 & 78.75 & \textbf{79.12} & 82.06 & 80.92 & \textbf{82.79}\\
\textbf{Mu} & 58.50 & 56.58 & 58.33 & 69.52 & 67.95 & \textbf{69.70} & 65.71 & 64.75 & \textbf{66.65} & 68.94 & 67.47 & \textbf{70.03} \\
\textbf{Pl} & 68.81 & \textbf{69.64} & 68.16 & 80.45 & \textbf{80.78} & 79.62 & 74.43 & \textbf{75.08} & \textbf{77.89} & 75.90 & \textbf{76.16} & \textbf{77.09} \\
\textbf{Bo} & 82.41 & 81.95 & \textbf{82.90} & 91.03 & 89.99 & \textbf{91.53} & 88.38 & 87.31 & \textbf{89.65} & 89.10 & 88.16 & \textbf{90.94} \\
\textbf{Se} & 75.88 & \textbf{77.23} & 74.45 & 86.41 & 86.35 & 86.95  & 86.96 & \textbf{87.37} & \textbf{87.70} & 88.33 & 88.08 & \textbf{88.48} \\
\textbf{Re} & 73.17 & 71.64 & 72.79 & 80.75 & 78.21 & \textbf{82.49} & 77.06 & 74.95 & \textbf{78.00} & 79.32 & 76.90 & \textbf{80.31} \\
\textbf{Cr} & 70.27 & \textbf{79.57} & 70.77 & 75.95 & \textbf{81.07} & 76.61 & 80.82 & \textbf{84.91} & 77.31  & 77.37 & \textbf{82.02} & 75.88 \\
\hline
\textbf{Avg.} & 71.73 & \textbf{72.70} & 71.32 & 80.91 & 80.78 & \textbf{81.24}  & 78.92 & 79.02 & \textbf{79.96} & 80.15 & 79.96 & \textbf{80.79} \\
\bottomrule[1pt]
\end{tabular}
\end{adjustbox}
\caption{Ablation Study of \textit{adaption-from-the-memory} and \textit{learn-from-the-memory} on 1-shot, 5-shot, 10-shot and 20-shot respectively on SNIPS dataset. B, A, and M stand for the strongest baseline \textit{L-ProtoNet+CDT+VPB}, only adaption-from-memory, and only learn-from-memory respectively.
}
\label{tab:aba_study}
\end{table*}

\section{Ablation Study and Analysis}

\subsection{Ablation Result}

We borrow the result from \citet{zhu2020vector} as baseline (i.e. L-ProtoNet+CDT+VPB) here since it reaches the best performance out of all baselines. Table~\ref{tab:aba_study} shows the ablation study of learning and adaption from memory. Comparing the result between 1-shot, 5-shot, 10-shot, and 20-shot, we find that the \textit{learn-from-the-memory} (i.e. M) module gets more important as the number of shots increases. We attribute this phenomenon to the more transferable representations due to more labeled data brought by more shots. However, the \textit{adaption-from-the-memory} cannot keep consistent improvement, we think this is caused by noise introduced by the adaption layer. After combining these two modules, the model can reach the best performance as reported in Table~\ref{tab:snip_main} and Table~\ref{tab:ner_main}. Compared with the strongest baseline, the averaged F1 score further improved (More analysis can be found in Appendix~\ref{appendix_ana}).

\subsection{t-SNE Visualization Analysis}
\begin{figure}[t]
\centering
\includegraphics[trim={0.2cm 0cm 0cm 0cm}, clip, width=0.45\textwidth]{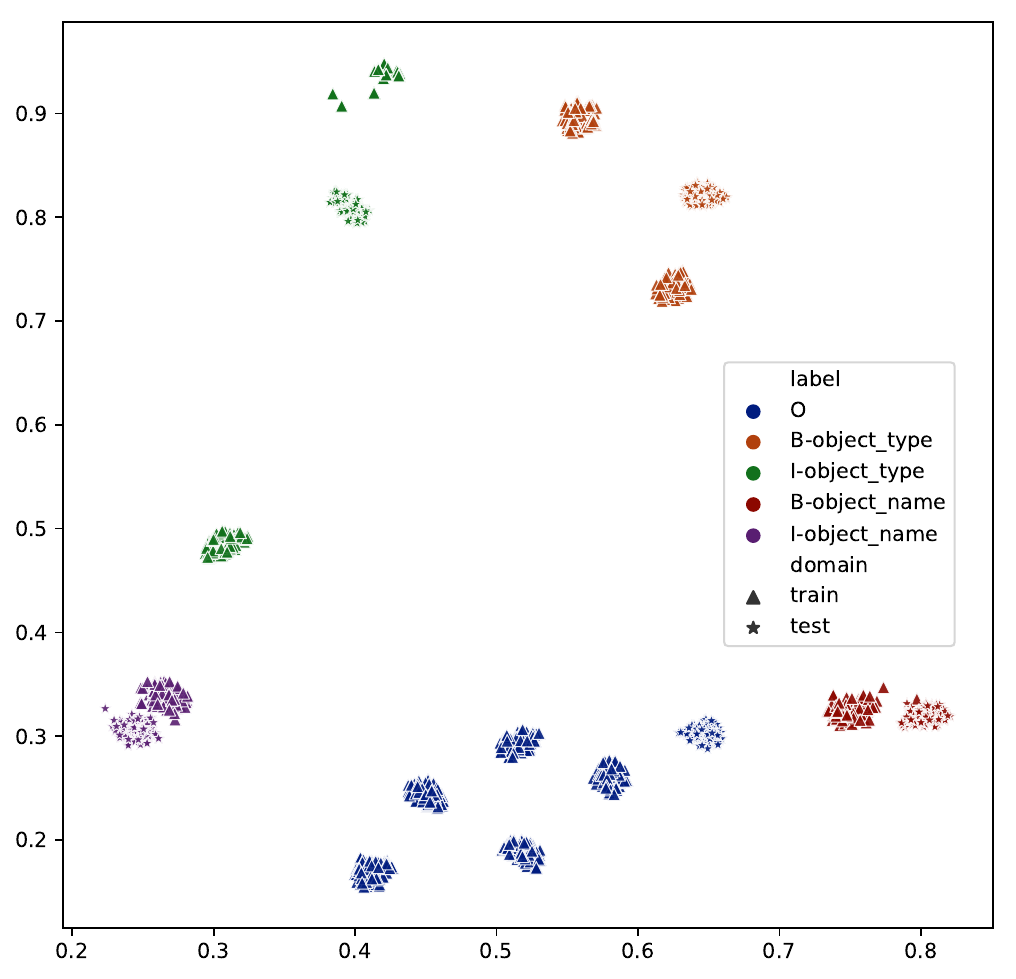}
\caption{The t-SNE visualization of label representations during meta-training and meta-testing learned by Metric-based Meta-learning Model (Target Domain: Cr 5-shot).}
\label{fig:origin_feature}
\end{figure}

\begin{figure}[t]
\centering
\includegraphics[trim={0cm 1cm 0cm 2cm}, clip, width=0.52\textwidth]{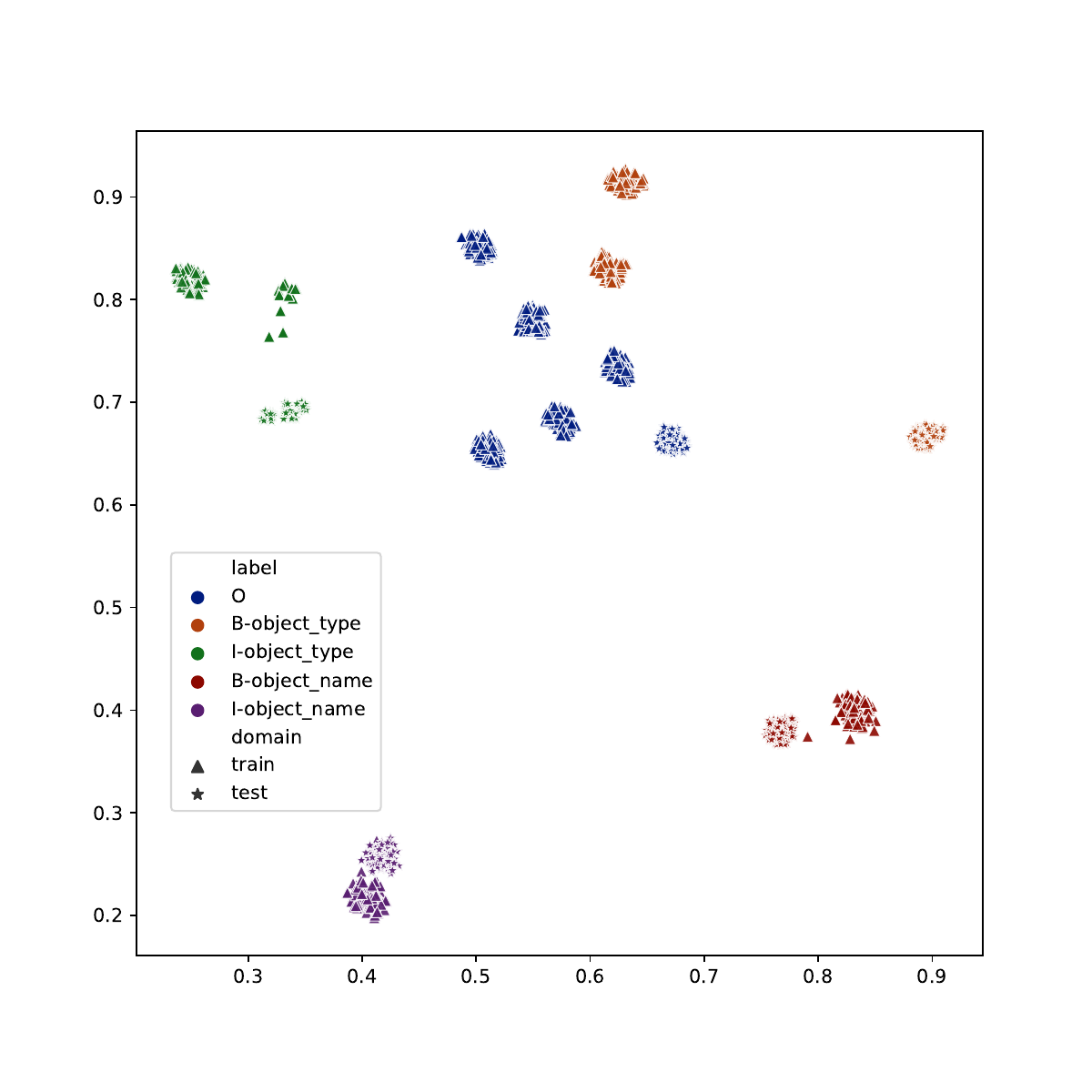}
\caption{The t-SNE visualization of label representations with our proposed learn-from-the-memory (Target Domain: Cr 5-shot).}
\label{fig:features}
\end{figure}

We present a \textit{t-SNE} visualization of label representations of trained metric-based meta-learning methods as shown in Figure~\ref{fig:origin_feature} and we additionally draw the \textit{t-SNE} visualization of label representations after adding contrastive learning constraint in Figure~\ref{fig:features}. On the one hand, it is observed from Figure~\ref{fig:origin_feature} that: 1) the representations of \textit{B-object\_type} and \textit{I-object\_type} at the meta-training stage are separated into distant groups; and 2) the representations at the meta-testing stage are shifted compared with those at the meta-training stage. For the first observation, we can conclude that the model can not remember what it already learned, failing to capture a consistent representation of the same label. A similar problem still happens at the meta-testing stage due to the presence of poorly sampled shots \cite{fei2021melr}. On the other hand, in Figure ~\ref{fig:features}, it is found that the distance between the representations of \textit{B-object\_type} (also \textit{I-object\_type}) during the meta-training stage is much closer, which proves the effectiveness of learn-from-the-memory to alleviate the sample forgetting problem. 

\begin{figure*}[h!]
    \centering
    \subfigure[1-shot]{%
    \label{fig:first}%
    \includegraphics[height=2.0in,width=0.45\textwidth]{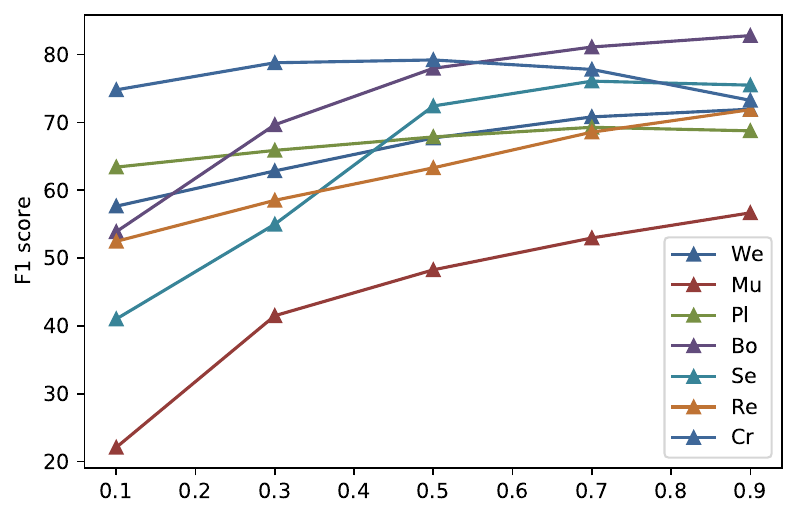}}%
    \qquad
    \subfigure[5-shot]{%
    \label{fig:second}%
    \includegraphics[height=2.0in,width=0.45\textwidth]{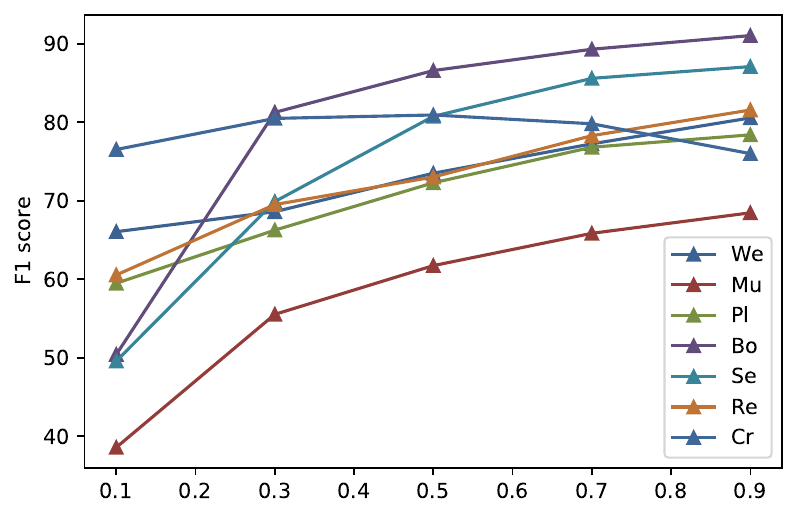}}%
    \caption{The impact of different scales on the performance, we suggest zooming in to check more details.}
    \vspace{-6mm}
    \label{fig:aba_scale}
\end{figure*}

\subsection{The Impact of different value of scale}
To investigate the effects of $\alpha$ during adaption-from-the-memory, we report the performance of different assignments of this scale. The result can be found in Figure~\ref{fig:aba_scale}. Since the larger the value, the more information from the meta-testing space, the less information from adaption. Thus, as long as the graph is monotonically increasing, the less useful the adaption is. However, as we can observe, it is obvious that not all domains show this trend. Specifically, under the 1-shot setting, "Pl", "Se" and "Cr" gets higher performance because of the adaption, and "Pl" and "Cr" continue this trend in the 5-shot. This shows the adaption layer is highly domain-sensitive and prefers the domain which has more overlapped labels. More analysis can be found in Appendix~\ref{appendix_overlap}.

\section{Conclusion}
In this paper, we address the \textit{sample forgetting problem} during meta-training and meta-testing stages in the metric-based meta-learning framework by capturing more transferable and informative label representations. To this end, we propose the Memory-based Contrastive Meta-learning (MCML) method, which consists of two modules: \textit{learn-from-the-memory} and \textit{adaption-from-the-memory} to function at different stages. Experimental results on both NER and SNIPS datasets demonstrate the advantages of our MCML framework in terms of scalability and robustness.



\section*{Limitations}
This paper tackles the issues of the sample forgetting problem in the metric-based meta-learning framework. We mainly focus on the few-shot slot tagging tasks but our proposed method is motivated by the unique setting of metric-based meta-learning which can be applied to other text classification tasks such as intent detection or news classification. We left this in our future work.

\section*{Acknowledgement}
We thank all reviewers for their insightful comments and suggestions. This research work is partially supported by ITF Project No. PRP/054/21FX and CUHK under Project No. 3230366.

\bibliography{custom}
\bibliographystyle{acl_natbib}
\clearpage
\appendix






\section{Analysis}
\label{appendix_ana}


\subsection{Less-shot or More-shot ?}
\label{appendix_less_or_more}

Table~\ref{tab:aba_study} shows the result of 10-shot and 20-shot on the SNIPS dataset which is generated following the method proposed by \citet{hou2020fewshot}.

\noindent \textbf{More Shots.} Compare 10-shot with 20-shot, we can find that all domains are improved with the help of \textit{learn-from-the-memory} when the number of shots increases except ``SearchCreativeWork". Since this is the only domain that has 100\% overlap labels during meta-training and meta-testing, we attribute this phenomenon caused by poor representations from meta-testing without \textit{adaption-from-memory}. 

\noindent \textbf{Fewer Shot v.s More Shot.} Compare 1-shot and 5-shot (less-shot) with 10-shot and 20-shot (more-shot), there are some interesting findings: 1) \textit{learn-from-the-memory} can boost 6 out of 7 domains in more-shot instead of 3 in less-shot. This demonstrates the importance and effectiveness of this module when the number of shots gets more; 2) \textit{adaption-from-memory} shows exactly the same gains whether or not there are more shots. This is reasonable since the number of shots does not affect the number of labels, and also the accuracy of adaption. We conclude that \textit{learn-from-the-memory} is always worth trying, and \textit{adaption-from-the-memory} highly depends on a specific domain.

\subsection{The Impact of overlap}
\label{appendix_overlap}

\begin{figure}[h]
    \centering
    \includegraphics[scale=0.5]{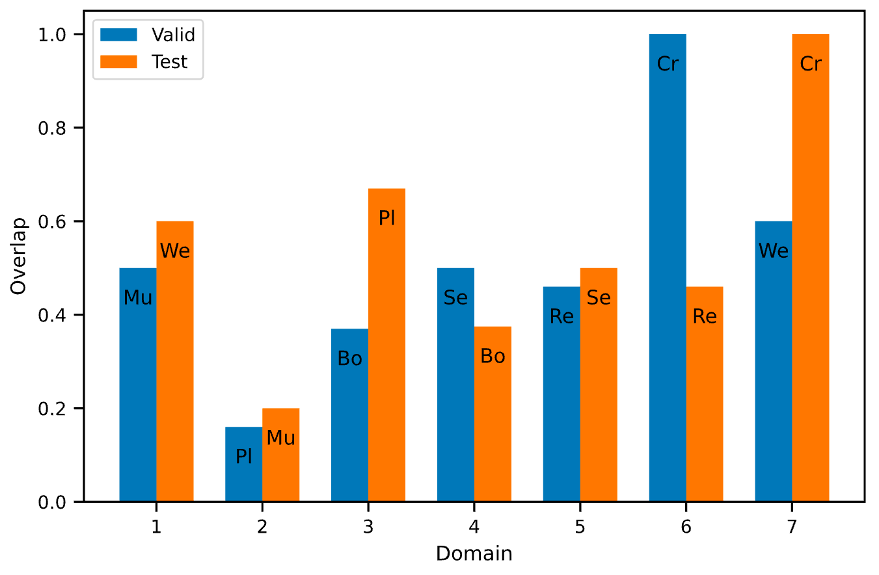}
    \caption{The Percentage of Overlap Labels between train and valid or test}
    \label{fig:overlap}
\end{figure}

The performance of the adaption function highly depends on the number of overlap labels. Since the more overlap labels between training and testing, we will get a more accurate adaption function. Figure~\ref{fig:overlap} shows the percentage of overlap labels between training data and validation or test data.

To further investigate to what extent the influence of overlap labels on adaption performance, we utilize the Pearson correlation coefficient to analyze the relationship between these two variables. The calculated result is 0.83 which shows these two variables are highly related.

\begin{figure}[h]
    \centering
    \includegraphics[scale=0.15]{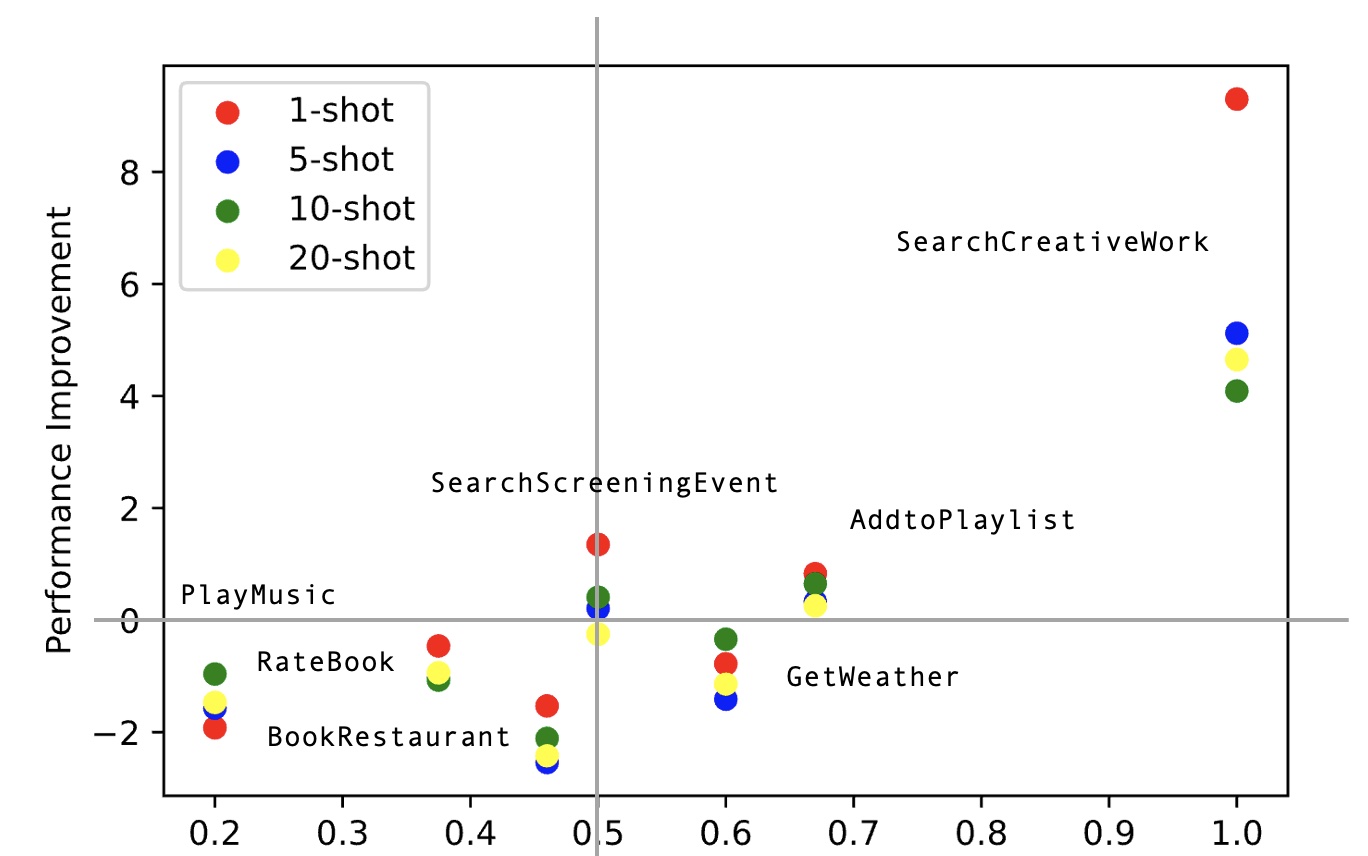}
    \caption{Improvement of different shots with different percentage overlapped labels.}
    \label{fig:overlap_imp}
\end{figure}

The performance improved by the adaption of different domains can be found in Figure ~\ref{fig:overlap_imp}. It is noted that although "GetWeather" domain has 60\% overlap labels with training data, the performance declines surprisingly. We further investigate the specific overlap labels of this domain, and we find most of them are "state", "country" and "city", common regular entity types which appear in almost every corpus. When the number of shots is less, these common entities cannot be represented accurately during meta-training, much less during adaption. This explains the poor performance of 1-shot and 5-shot and the higher performance of 10-shot and 20-shot. For the above reasons, we argue it is worth trying adaption once the overlap exceeds 50\% as long as the overlaps labels have some domain-specific features.

\end{document}